\documentclass[times]{elsarticle}

\usepackage{lineno,hyperref}

\journal{Remote Sensing of Environment}

\usepackage{multirow}
\usepackage{epstopdf}
\usepackage{url}
\usepackage{color}
\usepackage{array}
\usepackage{amsmath}
\usepackage{subcaption}

\begin{document}

\begin{frontmatter}

\title{A deep learning approach to detecting volcano deformation from satellite imagery using synthetic datasets}

\author[mymainaddress]{N. Anantrasirichai\corref{mycorrespondingauthor}}
\cortext[mycorrespondingauthor]{Corresponding author}
\ead{n.anantrasirichai@bristol.ac.uk}

\author[mysecondaryaddress]{J. Biggs}
\author[mysecondaryaddress]{F. Albino}
\author[mymainaddress]{D. Bull}

\address[mymainaddress]{Visual Information Laboratory, University of Bristol, UK}
\address[mysecondaryaddress]{School of Earth Sciences, University of Bristol, UK}

\begin{abstract}
Satellites enable widespread, regional or global surveillance of volcanoes and can provide the first indication of volcanic unrest or eruption. Here we consider Interferometric Synthetic Aperture Radar (InSAR), which can be employed to detect surface deformation with a strong statistical link to eruption. Recent developments in technology as well as improved computational power have resulted in unprecedented quantities of monitoring data, which can no longer be inspected manually. The ability of machine learning to automatically identify signals of interest in these large InSAR datasets has already been demonstrated, but data-driven techniques, such as convolutional neutral networks (CNN) require balanced training datasets of positive and negative signals to effectively differentiate between real deformation and noise. As only a small proportion of volcanoes are deforming and atmospheric noise is ubiquitous, the use of machine learning for detecting volcanic unrest is more challenging than many other applications. In this paper, we address this problem using synthetic interferograms to train the AlexNet CNN. The synthetic interferograms are composed of 3 parts: 1) deformation patterns based on a Monte Carlo selection of parameters for analytic forward models, 2) stratified atmospheric effects derived from weather models and 3) turbulent atmospheric effects based on statistical simulations of correlated noise. The AlexNet architecture trained with synthetic data outperforms that trained using real interferograms alone, based on classification accuracy and positive predictive value (PPV). However, the models used to generate the synthetic signals are a simplification of the natural processes, so we retrain the CNN with a combined dataset consisting of synthetic models and selected real examples, achieving a final PPV of 82\%. Although applying atmospheric corrections to the entire dataset is computationally expensive, it is relatively simple to apply them to the small subset of positive results. This further improves the detection performance without a significant increase in computational burden (PPV of 100\%). Thus, we demonstrate that training with synthetic examples can improve the ability of CNNs to detect volcano deformation in satellite images, and propose an efficient workflow for the development of automated systems. 
\end{abstract}


\end{frontmatter}


\section{Introduction}

Interferometric Synthetic Aperture Radar (InSAR) employs differences in the phase of radar waves returning to the satellite to generate maps of surface deformation.  Statistically the
deformation at volcanoes is statistically linked to eruption \citep{Biggs:Global:2014}, and unlike other satellite methods, is dominantly detected prior to eruption \cite{Furtney:Synthesizing:2018}. This could allow volcanologists to monitor volcanic activity in large and remote areas, which is particularly valuable in developing countries where expertise and ground monitoring equipment may be insufficient.
Contemporary satellites, such as Sentinel-1, provide global coverage, shorter timespan and high resolution images. This results in very large amount of data that makes manual inspection infeasible. 

InSAR images, known as interferograms, contain contributions both from volcanic deformation and the radar path through the atmosphere. The atmospheric artefacts can dwarf the deformation signal \cite{Parker:Systematic:2015,Ebmeier:Applicability:2013,pinel2014volcanology}, making simple threshold-based approaches to automatic detection impractical. Atmospheric corrections can be applied based on external data sources such weather models, or GPS tropospheric delays, or by applying statistical approaches to phase-elevation correlations or time-series \citep[e.g.][]{bekaert2015statistical,li2005interferometric,jolivet2014improving}. However, these are time consuming to apply to large datasets, and typically cannot be applied in real-time or to wrapped data.  Blind source separation techniques, such as Independent Component Analysis (ICA), have the potential to automatically isolate different signals, thus making changes in the deformation component easier to detect \cite{ebmeier2016, Gaddes2018}, but have so far only been tested on a few case studies. 

Deep convolutional neural networks (CNN) -- a class of neural networks inspired by deeply complex hierarchical structure of neurons that connect in multiple layers via learnable filters \cite{Krizhevsky:ImageNet:2012} -- is {one of the feasible methods} for automatically analysing global datasets. Our previous 'proof-of-concept' study demonstrated the ability of CNNs to detect rapidly deforming systems that generate multiple fringes in wrapped interferograms \cite{Anantrasirichai:Application:2018} but could not reliably distinguish between deformation signals and atmospheric artefacts in a small percentage of cases. Our approach is to use machine learning to interrogate the large dataset of wrapped interferograms and identify a subset of images to apply unwrapping algorithms and atmospheric corrections. Improving the efficiency of the algorithm by reducing the number of false positives will thus reduce the need for unwrapping and atmospheric correction. 

CNNs require a balanced dataset for training otherwise the algorithms can become 'over-tuned' to specific case studies \citep{Yan:deep:2015}. This is a challenge for this application because  few automatically-generated interferograms contain significant deformation signals - most are short-duration and cover volcanoes that are not deforming, or are deforming slowly. For example, Anantrasirichai et. al. \citep{Anantrasirichai:Application:2018} used a dataset of $>$30,000 Sentinel-1 interferograms produced by the LICSAR system which covered $\sim$900 volcanoes globally, but only contains 42 interferograms that show deformation signals. The imbalance in training data can be mitigated by artificially subsampling or upsampling the training set. Anantrasirichai et. al. \citep{Anantrasirichai:Application:2018} used programmatic data augmentation to increase the number of deformation samples by applying transformations (i.e. rotations, flips, distortions and pixel shifts) on the existing positive samples. However, the problem of a within-class imbalance is still present \citep{Japkowicz:concept:2001} because the characteristics of global volcanic deformation cannot be generalised using the limited number of deformation samples, even when augmented.

In this paper, we aim to improve the ability of the CNN model to distinguish deformation signals from atmospheric artefacts by using synthetic data to overcome imbalanced training data problem. The synthetic interferograms are generated from three main components, which are surface deformation, stratified atmosphere and turbulent atmosphere. The synthetic deformation signals are produced using simple elastic sources for earthquakes, dykes, sills and point pressure changes at magma chambers \citep{Mogi:Relation:1958, Okada:surface:1985,Fialko:Deformation:2001}. The stratified atmospheric interferograms are obtained from the Generic Atmospheric Correction Online Service (GACOS) \citep{yu2018interferometric}, whilst the turbulent atmospheric interferograms are simulated using the statistical characteristics of correlated noise in real interferograms \citep{Lohman:noise:2005, biggs2007multi}. The classification method is developed  through  a transfer learning strategy by fine-tuning a pretrained CNN network.

\section{Convolutional Neural Networks and Training Dataset Problems}
\label{sec:cnn}


Machine learning (ML) is a popular approach to data analysis that automatically discriminates input patterns into learnt or defined classes. The most popular, and perhaps most powerful, ML tools for image classification and recognition are deep convolution neural networks (CNNs). These data-driven approaches are hierarchical feature learning methods, which are straightforward to adapt to most specific applications without the need for manual feature extraction. However, the main drawback of CNNs is that the most efficient and successful models require a large training set of labelled data.  

\subsection{Convolutional Neural Networks}
\label{ssec:cnn}

Convolutional neural networks (CNNs) are a class of deep feed-forward artificial neural networks. They comprise a series of convolutional layers that are designed to take advantage of 2D structures, such as an image. These convolutional layers employ locally connected layers that apply convolution between a predefined-size kernel and an internal signal, i.e. the output of the convolutional layer is the input signal modified by a filter. The weights of the filter are adjusted using a loss function and backpropagation (the backward propagation of errors) through multiple forward and backward iterations. This aims to determine what features are being detected associating to nature of the training data. The early layers extract low-level features conceptually similar to vision basis functions found in the primary visual cortex \citep{Matsugu:Subject:2003}.  

The most common architecture of a CNN has the convolution layer connected to a pooling layer, which combines the outputs of neuron clusters at one layer into a single neuron. {Some architectures omit pooling layers to obtain dense features \cite{Li:CSRNet:2018, Anantrasirichai:DefectNET:2019}.} Subsequently, activation functions such as {\textit{tanh}} (the hyperbolic tangent) or {\textit{ReLU}} (Rectified Linear Unit) are applied to introduce non-linearity into the networks \citep{Agostinelli:Learning:2015}. This structure is repeated with similar or different kernel sizes. As a result, the CNN learns to detect edges from the raw pixels in the first layer, then uses the edges to detect simple shapes in the next layer, for example. The higher layers produce higher-level features, which have more semantic meanings. The last few layers are the classification part. It consists of some fully connected layers, having full connections to all the activations in the previous layer, and a softmax layer, where the output class is modelled as a probability distribution - exponentially scaling the output to be between 0 and 1 (also called normalised exponential function).

\subsection{Imbalanced training data and solutions}
\label{ssec:imbalance}

Imbalanced data problems in classification occur when data sets have skewed class distributions, i.e. the majority of data instances belong to one class and far fewer instances belong to others. This causes classifier algorithms to have a bias towards classes which have a greater number of instances and preferentially predict majority class data. Features of the minority class are treated as noise and are often ignored. Although deep CNN approaches often perform better than traditional machine learning methods in many applications, their performance can be worse with imbalanced datasets \citep{Yan:deep:2015}.

Numerous approaches have been introduced to create balanced distributions of data and these can be divided into two major groups: modification of the learning algorithm, and data manipulation techniques \citep{He:Learning:2009}. The first group modifies existing algorithms to give greater emphasis to the minority classes. This can be achieved using cost-sensitive learning which  assigns costs with a higher penalty for minority class samples \citep{ThaiNghe:cost:2010}. A cost matrix representing the cost of each type of misclassification is applied and the result is the class with minimum expected cost, described by the summation of all class probability estimations weighted with the cost matrix. However, it is generally difficult to optimise the cost matrix for this method. 
The second group tries to rebalance the class distributions of the training data. Typical methods include downsampling majority classes, oversampling minority classes, or both.
This group is favoured and simpler as the only change needed is the training data rather than the learning algorithms. The disadvantage of downsampling is that many data instances in the majority class are ignored, which may result in the loss of information.
In contrast, the synthetic minority oversampling technique (SMOTE) is a powerful method that creates synthetic data points from the existing ones \citep{Chawla:SMOTE:2002}.

However, none of these approaches are suitable for the problem of deformation classification in InSAR datasets. The number of interferograms showing deformation is approximately 0.15\% of all acquired interferograms, and if computing in pixels, the ratio of positive and negative areas is only 1:15,000.  The global dataset used in \citep{Anantrasirichai:Application:2018} covers over 900 volcanoes in 2016-2017, but only 4 volcanoes deformed, namely Cerro Azul, Sierra Negra, Etna and Erta Ale. This means that all existing methods would likely lead to overfitting as the characteristics of known ground deformation are not generic enough to represent deformation at the wide range of volcanoes globally. Therefore, we propose generating synthetic data to improve classification performance,  using established models that represent existing data well, but are flexible enough to generate a wider range of possible signals.

\section{Generation of Synthetic Training Data}
\label{sec:Generation}

CNNs are data-driven methods, so it is critical to train the networks using appropriate data. In this paper, we use existing models to create synthetic examples of a) deformation, b) stratified atmospheric artefacts and c) turbulent atmospheric artefacts.  We use a Monte Carlo approach to select parameter values, thereby including scenarios that are considered feasible but have not actually been observed.  The resulting synthetic training datasets should provide a better generalisation than the real dataset.

InSAR produces maps of phase change between two time-separated radar images. The phase shift is a combination of i) satellite viewing geometry, ii)  instrument thermal noise, iii) atmospheric delay, iv) systematic changes to dielectric properties, v) scattering properties of a pixel, and vi) surface deformation  \citep[e.g.][]{Burgmann:Synthetic:2000,biggs2007multi,Ebmeier:Applicability:2013}. The atmospheric delay can be decomposed into atmospheric stratification and turbulent mixing \citep{Hanssen:radar:2001}. The first component results in phase delays correlated with topography, whilst the second component is frequently considered as random patterns in space and time, with spatial correlation over distances of $\sim$10~km \citep{Hanssen:radar:2001}.

\begin{figure*}[!ht]
	\centering
      		\includegraphics[width=\textwidth]{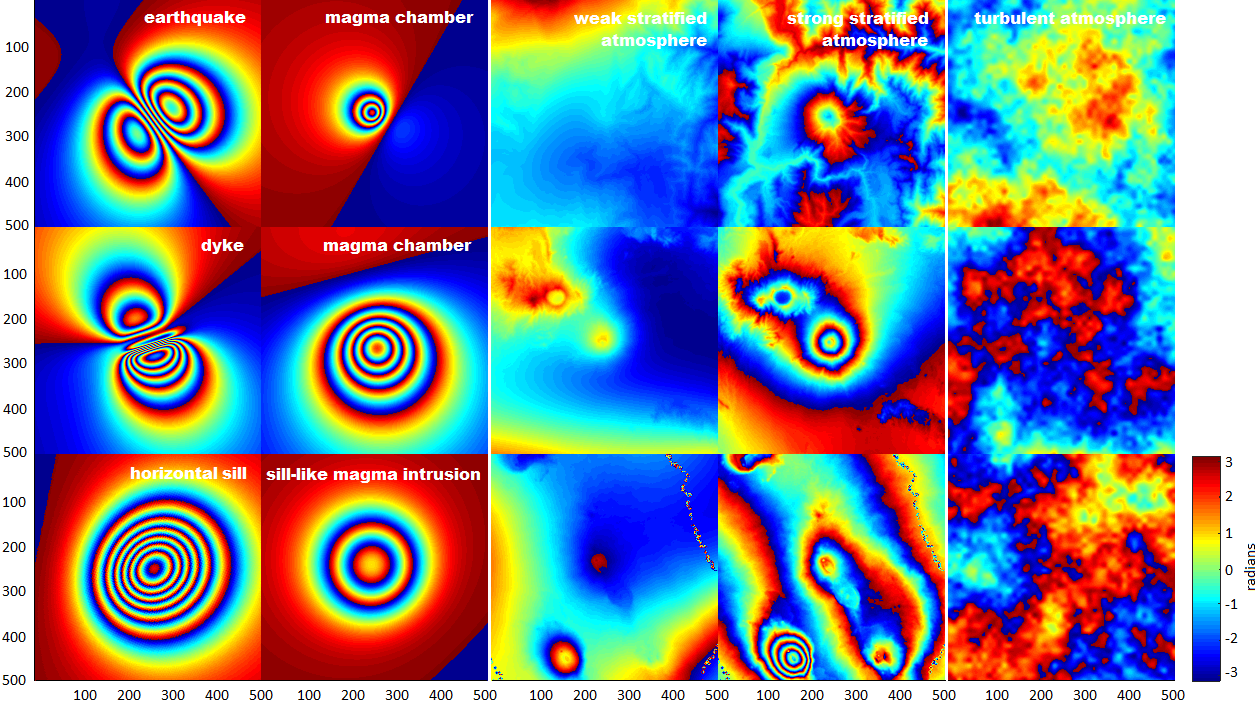}
					\caption{Synthetic components used for generating synthetic interferograms  (shown in wrapped angles in radians with the size of 500$\times$500 pixels).  Column 1 and 2 show different types of deformations. Column 3 and 4 show weak and strong stratified atmospheres (using the Generic Atmospheric Correction Online Service (GACOS)) obtained from the same locations, which are Tungurahua, San Miguel and Erta Ale from top to bottom rows, respectively. Column 5 shows turbulent atmospheres from low to high $\sigma^2_{max}$ in the top to the bottom rows, respectively. }
    \label{fig:synexamplesRGB}
\end{figure*}

We generate the 10,000 synthetic images for each of the 3 components under consideration, namely deformation $D$, stratified atmosphere $S$, and turbulent atmosphere $T$. Each image represents a region spanning $\sim$0.5$^\circ$ in latitude and longitude (equivalent to an image resolution of 500$\times$500 pixels for the Sentinel-1 dataset). Fig. \ref{fig:synexamplesRGB} shows example synthetic images of $D$, $S$ and $T$ (converted to wrapped images). The methods of synthetic data generation are described as follows.

\subsection{Deformation}
\label{ssec:Deformation}

We synthesise deformation signals, $D$, using widely-used analytic solutions for describing the surface deformation associated with simple geometric sources embedded in an elastic half-space. We used a Monte Carlo approach to select source parameters and project the 3-D surface displacement into the satellite line-of-sight using incidence angles of 0$^\circ$-45$^\circ$, and heading angles of 0$^\circ$-360$^\circ$. The CNN is not sensitive to orientation, so this represents the widest range of angles between source and viewing geometries.   Inflation and deflation of a magma chamber is modelled using a point pressure source (Mogi) model \citep{Mogi:Relation:1958}, with depths of 1-10~km, and volumes of 10$^5$-10$^7$~m$^3$. For sill-like magma intrusions, the displacement is calculated using a model of a horizontal circular (penny shaped) crack \citep{Fialko:Deformation:2001} using a radius of 0.5-6~km, pressure changes of 10$^5$-10$^7$~Pa and depths of 0.5-6~km$^3$.   Deformation due to earthquakes, dykes and horizontal sills is modelled using an Okada dislocation model \citep{Okada:surface:1985}, which describes shear and tensile dislocations.  Earthquakes are allocated strikes in the range 0$^{\circ}$-360$^\circ$, dip of 45$^{\circ}$-90$^\circ$, rake of 0$^{\circ}$-360$^\circ$, length of 0.5-10~km, depth  of 1-15~km, and slip of 0.5-2~m. Dykes are allocated strikes in the range 0$^{\circ}$-360$^\circ$, dip of 45$^{\circ}$-90$^\circ$, length of 2-8~km, depth  of 1-5~km, and opening of $<1$~m. Horizontal sills are allocated strikes in the range 0$^{\circ}$-360$^\circ$, dip of 0$^{\circ}$-10$^\circ$, length and width up to 5~km and depth of up to 6~km.

\subsection{Stratified atmosphere}
\label{ssec:Stratified}

We use the Generic Atmospheric Correction Online Service (GACOS) to model stratified atmosphere, $S$, based on weather model data \citep{yu2017generation, yu2018interferometric, yu2018generic}. Zenith total delay (ZTD) maps are derived from the high-resolution water vapour delays (0.125$^\circ$ and 6-hour resolutions) generated by the European Centre for Medium-Range Weather Forecasts (ECMWF). GACOS uses an Iterative Tropospheric Decomposition (ITD) model \citep{yu2017generation} to separate stratified and turbulent signals from tropospheric total delays and the final ZTD maps are interpolated to 90m spatial resolution and the time of acquisition using the Digital Elevation Model (DEM) from the Shuttle Radar Topographic Mission (SRTM). We generated 100 GACOS tropospheric delay maps from each of 100 representative volcanoes with 12-day intervals between images starting from 1 January 2016. We account for the different between zenith and the satellite line of sight by applying a scalar factor representative of Sentinel-1 incidence angles.

\subsection{Turbulent atmosphere}
\label{ssec:Turbulent}

Turbulent atmospheric delays, $T$, are spatially correlated and their covariance can be described using an exponentially decaying  function. For simplicity, the statistical properties of the atmosphere are assumed to be radially symmetric and have a homogeneous structure across the interferogram \cite{Hanssen:radar:2001,Parsons:1994:2006}. The one-dimensional covariance function is $c_{ij} = \sigma^2_{max} e^{(-\kappa d_{ij})}$, where $c_{ij}$ is the covariance between pixels $i$ and $j$, $d_{ij}$ is the distance between the pixels, $\sigma^2_{max}$ is maximum covariance and $\kappa$ is the decay constant, which is equivalent to the inverse of the $e$-folding wavelength \cite{biggs2007multi,Parker:Investigating:2014}. We can estimate these parameters from real interferograms, and based on all available 30,249 Sentinel-1 interferograms, we employ $\sigma^2_{max}$ of 5 - 9~mm$^2$ and $\kappa$ of 4 - 18~km. We use Monte Carlo samples of these distributions to generate synthetic variance-covariance matrices and use a Cholesky decomposition to produce synthetic images with the corresponding statistical properties.

\section{Method Development}
\label{sec:proposed}

\begin{figure*}[!ht]
	\centering
      		\includegraphics[width=\linewidth]{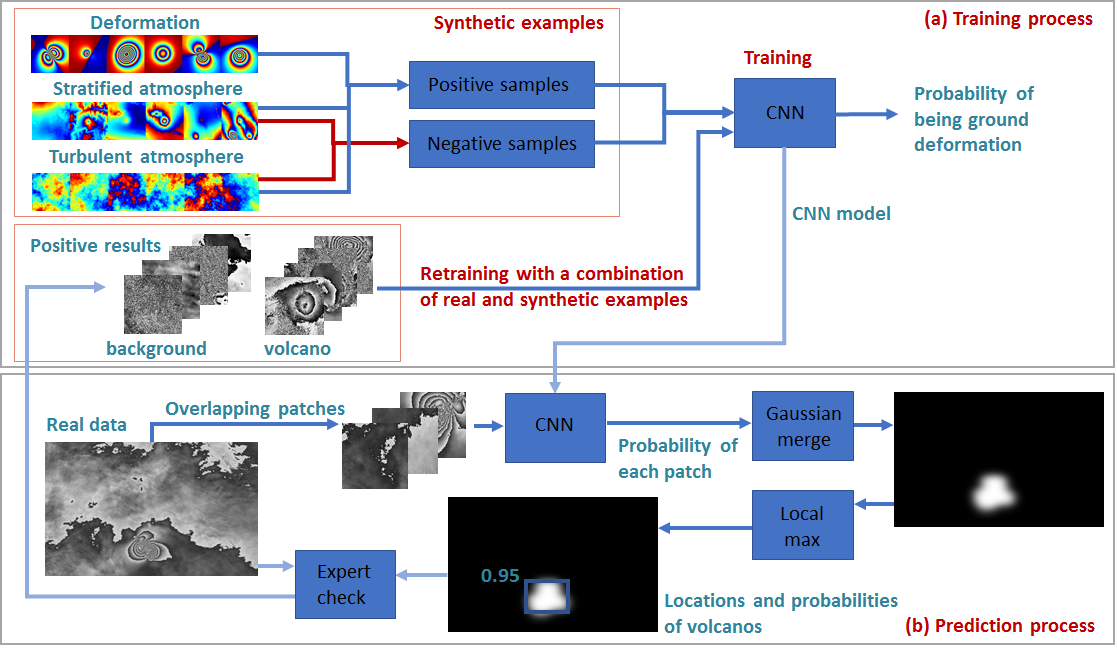}
					\caption{The proposed framework consists of two parts: (a) the training process and (b) the prediction process. {For training, synthetic examples are first employed to train the CNN to obtain the initial model. The prediction process tests the patches of new interferograms and gives the outputs as the probabilities $P$ of being ground deformation, which are merged with Gaussian weights.}  Finally, the expert checks the result, and the true and false positives are included to retrain the CNN using a combination of real and synthetic examples for better performance. CNN = convolutional neutral network.}
    \label{fig:workdiagram}
\end{figure*}

The proposed machine learning framework shown in Figure \ref{fig:workdiagram} employs convolutional neural networks to identify volcanic deformation in InSAR data. Initially we train the network using just the synthetic images (see section \ref{ssec:intialmodel}). {They are labelled as 1 or positive, where deformation is included; and 0 or negative in other combinations. Then, the initial model is employed in the prediction process (see section \ref{ssec:testrealdata}), where the new interferogram is divided into overlapping patches and those containing phase jumps are tested with the trained CNN model.} The results are then checked by an expert, the model is retrained, and the classification is repeated (see section \ref{ssec:retraining}). 

\subsection{Initial models with synthetic data}
\label{ssec:intialmodel}

 We use a transfer learning strategy, which involves fine-tuning a pretrained network rather than training a new network by initialising weights and biases with zeros or random values. The training process of this approach is faster as the parameters and features of the pretrained networks have been learnt using a large number and a variety of natural images, which can be classified up to 1,000 categories. In this paper, we aim to classify two categories, volcanic deformation and non-deformation, so the last two layers -— a fully connected layer and a softmax layer -— are amended and they are learnt with significantly faster learning rate than other layers that are directly transferred from the pretrained network. We use AlexNet as our previous study demonstrated that it outperforms other pretrained networks for this application \citep{Anantrasirichai:Application:2018}. {AlexNet contains five convolutional layers and three fully connected layers. The first, the second and the fifth convolutional layers are followed by max-pooling layers. ReLU is applied after very convolutional and fully connected layer. 
 Our previous work found that the validation accuracy saturates around epoch 30-40 (one epoch is when the entire dataset is passed forward and backward through the neural network), so we set the maximum number of epochs to 50.  {Setting the number of epochs too high could result in over-fitting, and early stopping can be used to stop the training process when validation errors start to increase \cite{Goodfellow-et-al-2016}.} The entire dataset cannot be fed into the neural network at once, so it is divided into multiple batches. We set the batch size to the maximum of the available system memory which is 100}. The output of the softmax layer is the predicted probability for each class. The symbol $P$ in this paper represents the probability of the interferogram containing a component of deformation.

\begin{figure*}[!ht]
	\centering
		\includegraphics[width=\linewidth]{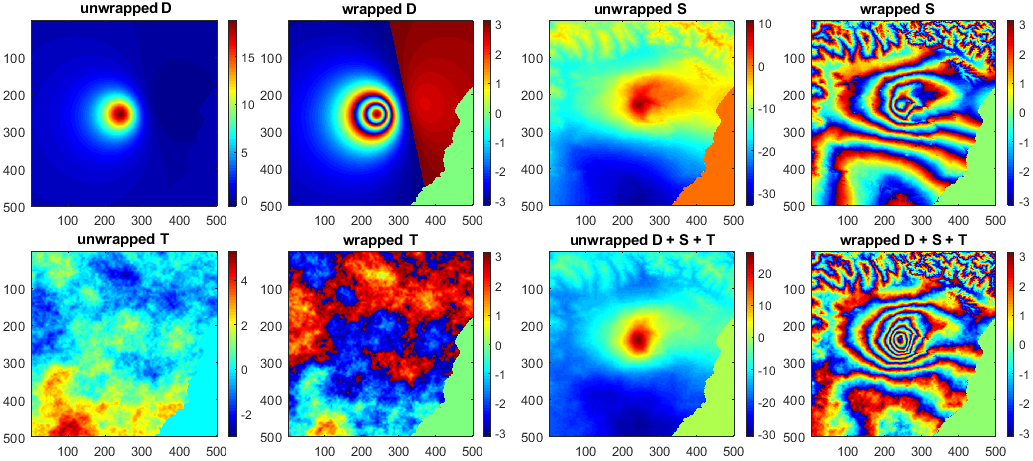}
		A
      	\includegraphics[width=\linewidth]{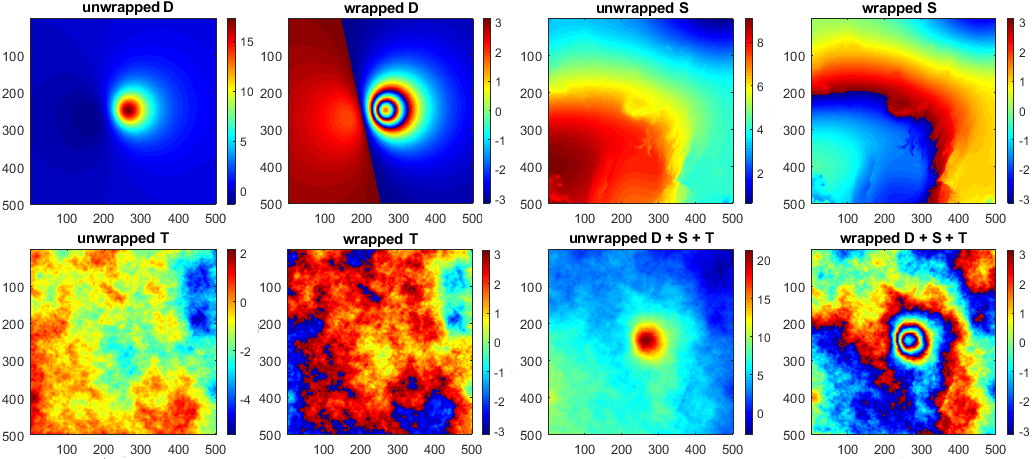}
      	B
	\caption{Two examples  of synthetic interferograms created by combining synthetic unwrapped deformation ($D$), stratified atmosphere ($S$) and turbulent atmosphere ($T$). 
{Both example A and B use $D$ due to a volcanic magma chamber.  A is generated with strong turbulent and stratified atmospheric conditions ($S$ is taken from Etna, 20150805-20150817), whilst B employs small atmospheric artefact ($S$ is taken from Alutu, 20150101-20150125).
 The parameters for generating $T$ are estimated from the real interferograms of the same areas. The unit of the colourbar is radian. }}
    \label{fig:syndata}
\end{figure*} 

We use the 10,000 synthetic examples of each component ($D$, $S$, and $T$) to create synthetic interferograms which are the summation of two or three signals with equal or unequal weights, i.e. $aD$+$bS$+$cT$, where $a,b,c \in [0,1]$. Two examples of synthetic interferograms are shown in Fig. \ref{fig:syndata}, {including a difficult case, A, and an easy case, B. Both example A and B use $D$ due to a volcanic magma chamber with volume change of 10$^7$~m$^3$ at depth of 5 km and incident angle of 30$^\circ$, but opposing heading angles. A is an example of the challenges of classification due to strong turbulent and stratified atmospheric conditions. In contrast,  B is an easier case as the topographically-correlated atmospheric artefact, $S$, is small.}
The synthetic interferograms are then cropped to the input size for the CNN (e.g. 224$\times$224 pixels for AlexNet \citep{Krizhevsky:ImageNet:2012}) and wrapped to the interval $[-\pi,\pi]$. For the purposes of machine learning, we convert the wrapped interferograms into grayscale images, i.e. the pixel value in the range of $[-\pi, \pi]$ is scaled to $[0,255]$ or $[-125, 125]$ if zero-centre normalisation is required. For the following combinations, each class contains 10,000 synthetic wrapped interferograms.

\begin{enumerate}
	\item \textbf{2-class model}: The model is trained with 2 classes: deformation and non deformation. We generated the training data by combining signals $D+S+T$ for the deformation class and only $S+T$ for the non deformation class. For each combined signal, the components $D$, $S$ and $T$ are randomly selected. 
	\item \textbf{3-class model}: Initially we trained the CNN with completely separate $D$, $S$ and $T$ signals, but this is a poor representation of real datasets, so we also trained the classifier with several more realistic combinations (e.g. $D+S,D+T, D+S+T$) as shown in Table \ref{tab:Classification}.
	\item \textbf{91-class model}: 
We generated weighted interferograms ($I$) by combining three components as $I = (\alpha D+\beta S+ \gamma T)/(\alpha + \beta + \gamma)$, where $(\alpha,\beta,\gamma) \in [0,0.25,0.5,0.75,1]$. Varying three weights with five values  creates 91 unique combinations, resulting in 91 classes: class 1 is [$\alpha_1$=0, $\beta_1$=0, $\gamma_1$=1]; class 2 is [$\alpha_2$=0, $\beta_2$=0.25, $\gamma_2$=0.75]; class 3 is [$\alpha_3$=0, $\beta_3$=0.5, $\gamma_3$=0.5]; ...; {class 91 is [$\alpha_{91}$=1, $\beta_{91}$=1, $\gamma_{91}$=1]}. We then apply a weight estimation approach using a multinomial classification. We estimate the strength of each component as multi-class problem and the model outputs a probability of each weight for each class $P_c=\{P_{\alpha}, P_{\beta}, P_{\gamma} \}_c$. The final predicted weight $w_{final}=\{{\alpha}, {\beta}, {\gamma} \}_{final}$ is

\begin{equation}
\label{eqn:finalprob}
		w_{final} = \sum_{c=1}^{91} w_c P_c, \: w_c \in \{\alpha_c,\beta_c,\gamma_c\}
\end{equation}

\end{enumerate}

{The training processes were run on a graphics processing unit (GPU) at the High Performance Computing facility (BlueCrystal phase 4) at the University of Bristol. The 2-class, 3-class, and 91-class models  were completed in approximately 10, 14 hours and 108 hours, respectively.}
The results are shown in Table \ref{tab:Classification}, including classification accuracy (Acc.) and class recall (RC), where RC$_c$ is the recall of class $c$, numbered following the order of the model name, e.g. for model ``$D$+$S$ vs $S$ vs $T$", $c$=1 for class $D$+$S$, $c$=2 for class $S$, and $c$=3 for class $T$. All models perform well on synthetic testing data (the accuracy and class recall are all more than 90\%).

\begin{table*}
\caption{Classification performances of CNN models trained by synthetic data and a combination of real and synthetic examples. Each model was tested with both the synthetic testing data and the real data. For the synthetic testing data, the performance is evaluated with classification accuracy (Acc.) and class recall (RC), where RC$_c$ is the recall of class $c$. For the 91-class model, the performance of the prediction is measured with the mean square error (MSE). MSE$_{all}$ is the MSE of all predicted weights, and MSE$_{\alpha}$, MSE$_{\beta}$, MSE$_{\gamma}$ are the MSEs of predicted $\alpha$, $\beta$ and $\gamma$, respectively. For the real data, the results show the performance of deformation detection. This evaluated on 30,249 interferograms of the Sentinel-1 dataset, of which 42 interferograms were marked as true deformations. The objective results show the total number of predicted positives (P), the numbers of confirmed true positives (TP), confirmed false positives (FP), and confirmed false negatives (FN). }
 \centering
\scriptsize
 \begin{tabular}{ll|cccc|cccc }
 \hline
\multicolumn{2}{c}{\multirow{ 2}{*}{Model}} & \multicolumn{4}{c}{Synthetic testing data}  &	\multicolumn{4}{c}{Real data (Sentinel-1)} \\ \cline{3-10}
 & & Acc. & RC$_1$ & RC$_2$ & RC$_3$  &	\# P  & \# TP &	\# FP   & \# FN\\
 \hline
 \multirow{2}{*}{2-class} & Initial (Envisat) & 0.893 & 0.943 & 0.844 & - & 1369 & 42	& 1327	& 0 \\
& Combination & 0.897 & 0.880 & 0.974  & - & 104 & 42	& 62	& 0  \\
\hline
 \multirow{2}{*}{2-class} & $D$+$S$+$T$ vs $S$+$T$ & 0.981 & 0.986 & 0.975 & - & 363 & 41	& 321	& 1 \\
& Combination & 0.976 & 0.972 & 0.989  & - & 52 & 41	& 11	& 1 \\
\hline
 \multirow{6}{*}{3-class} & $D$ vs $S$ vs $T$ & 1.000 & 0.999 & 0.998 & 1.000	& 0 & 0		& 0	&41\\
	&	$D$+$S$ vs $S$ vs $T$ &  0.998 &  0.998 & 0.998 & 1.000 & 18 & 18 &  0  & 24\\
  &	$D$+$T$ vs $S$ vs $T$ & 0.979 & 0.944 & 0.993 & 1.000  & 1411 & 42	& 1369	 & 0	\\
  &	$D$+$S$+$T$ vs $S$ vs $T$ & 0.993 & 0.993 & 0.986 & 1.000 & 1370 & 42	& 1328	& 0	\\
	&	$D$+$S$+$T$ vs $S$+$T$ vs $T$ & 0.911 & 0.992 & 0.965 & 0.976 & 1160& 42	& 1118	 & 0	\\
	& Combination & 0.953 & 0.977 & 0.930  & 0.991 & 83 & 42	& 41	& 0 \\
	\hline
	\multirow{3}{*}{91-class} & \multirow{2}{*}{$\alpha D$+$\beta S$+$\gamma T$} & \multicolumn{4}{l}{MSE$_{all}$=0.156, MSE$_{\alpha}$=0.068} & \multirow{2}{*}{334} & \multirow{2}{*}{38} & \multirow{2}{*}{295} & \multirow{2}{*}{1} \\
	& & \multicolumn{4}{l}{MSE$_{\beta}$=0.099, MSE$_{\gamma}$=0.071} & & &  &\\
	& Combination & - & - & -  & - & 50 & 41	& 9	& 1 \\
\hline
 \end{tabular}
 \label{tab:Classification}
\end{table*}

\subsection{Testing with real data}
\label{ssec:testrealdata}
Next we investigate how well a CNN trained with synthetic signals performs on real interferograms using the same dataset as our proof-of-concept study \cite{Anantrasirichai:Application:2018}. The InSAR data was acquired by the Sentinel-1 radar mission operated by the European Space Agency (ESA), and processed with the automated InSAR processing system LiCSAR (http://comet.nerc.ac.uk/COMET-LiCS-portal/) developed by the Centre for Observation and Modelling of Earthquakes, Volcanoes and Tectonics (COMET). Our dataset consists of 30,249 interferograms covering $\sim$900 volcanoes during the year 2016 and 2017. The data set is weighted towards European volcanoes, which correspond to almost 50 \% of the total available images because the orbit cycle is the shortest (every 6 days) and the LiCSAR system has been running for the longest time period (2 years). The LiCSAR system routinely calculates inteferograms for the three closest combinations, forming a trio of interferograms of increasing time-span. Each interferogram is cropped to a region spanning 0.5$^{\circ}$ in latitude and longitude centered on the volcano edifice. From this global dataset, we expect to fully explore the range of InSAR atmospheric and deformation signals as the volcanoes studied are located in different climate environments (e.g. temperate, tropical and arid) and have different morphologies ranging from steep stratovolcanoes, to shield volcanoes or calderas.

During the prediction process, we divide the real interferogram into overlapping patches at the required input size for AlexNet (224$\times$224 pixels).  The top-left position of each patch is then repeatedly shifted by 28 pixels to cover the entire image. We then employ Canny edge detection \citep{Canny:Edge:1986} to detect where the phase jumps between -$\pi$ and $\pi$. The patches containing the phase discontinuities are fed to the trained CNN model to obtain the probability that they represent ground deformation. Homogeneous areas, where there are no strong edges associated with phase discontinuities, are unlikely to contain rapid volcanic deformation. These patches are hence instantly defined as background and are not tested by the CNN \cite{Anantrasirichai:Application:2018}. Finally the output probabilities from overlapping patches are merged using a rotationally symmetric Gaussian lowpass filter with a size of 20 pixels and standard deviation of 5 pixels. {The highest probability $P_{max}$ and its location are indicated.}
The performances shown in Table \ref{tab:Classification} are chosen to emphasise the ability to detect deformation. This means that if the probability of the class containing $D$ or the weight $\alpha$ is the largest, that interferogram is classified as deformation or positive (P). 

Table \ref{tab:Classification} shows the number of the Sentinel-1 interferograms correctly classified as containing deformation (true positive, TP), incorrectly (false positive, FP), and misdetection (false negative, FN). The experiments show that the cleaner signals (``$D$ vs $S$ vs $T$" and ``$D$+$S$ vs $S$ vs $T$") are easier to classify, but the models trained by them are not suitable for real interferograms. Combining the turbulent atmospheric $T$ and deformation $D$ components during training improves performance better than combining stratified atmospheric $S$ and deformation components $D$ (i.e. the model of ``$D$+$T$ vs $S$ vs $T$" performs better on real data than that of ``$D$+$S$ vs $S$ vs $T$"). However, the combined $D$+$S$+$T$ signals give the best performance as their characteristics are closest to the real interferograms.

\subsection{Retraining with a combination of real and synthetic examples}
\label{ssec:retraining}
{The CNN trained with only synthetic data outperforms the initial training using earlier Envisat data, which found 1327 false positives, but still generates a significant number of FP (321 for the 2-class model, $>$1000 for the 3-class model and 295 for the 91-class model). This is because the synthetic examples are a simplification of the real signals - for example, the deformation models only consider simplified source geometries and homogeneous elastic media, while the the turbulent models only consider radially symmetric conditions. Hence the synthetic data cannot fully reproduce the characteristics of the natural signals. We therefore retrain the algorithm with a combination of the synthetic models and some selected real examples.  

For the combined training dataset, we select real examples using the results of the previous algorithm.}
The data preparation is straightforward for the 2-class model: the patches included in class $D$+$S$+$T$ are i) those of the TP interferograms that have $P_{D+S+T}>0.5$  and ii) the deformation patches of the FN interferograms. The patches of the FP interferograms that have $P_{S+T}>0.5$  are used as class $S$+$T$. For the 3-class model, the model of ``$D$+$S$+$T$ vs $S$+$T$ vs $T$" gives the best performance and is selected for further analysis.  The patches included in class $D$+$S$+$T$ are i) those of the TP interferograms that have $P_{D+S+T}>1/3$, and ii) the deformation patches of the FN interferograms.  The patches of the FP interferograms that have $P_{S+T}>1/3$  are used as class $S$+$T$. For balancing the training samples, new synthetic $T$ signals are also added in class $T$ and these are generated using $\sigma^2_{max}$ and $\kappa$ computed from the FP interferograms. 

For 91-class model, the patches of the TP interferograms having $\alpha > \beta,\gamma$  and the deformation patches of FN interferograms are used as class [$\alpha$=1, $\beta$=0, $\gamma$=0]. For the FP interferograms, the turbulence parameter $\sigma^2_{j,max}$ is calculated from the patch $j$. It is then used to estimate the weight $\gamma_j$, which equals to $(\sigma^2_{j,max} - m)/r$, where $m$ and $r$ are the minimum and the range values of all $\sigma^2_{max}$ of synthetic $T$ (here, $m$=5 and $r$=$9-5$=4 as mentioned in Section \ref{ssec:Turbulent}). Consequently the patch $j$ is used in the class ${\tilde{c}}$, where [$\alpha_{\tilde{c}}$=0, $\beta$=1$-\gamma_{\tilde{c}}$, $\gamma$=$\gamma_{\tilde{c}}$] and $\tilde{c} = arg \underset{c}\min \{(1-\gamma_j-\beta_c)^2 + (\gamma_j-\gamma_c)^2\}$.

\begin{figure*}[!ht]
	\centering
      	\includegraphics[width=0.75\linewidth]{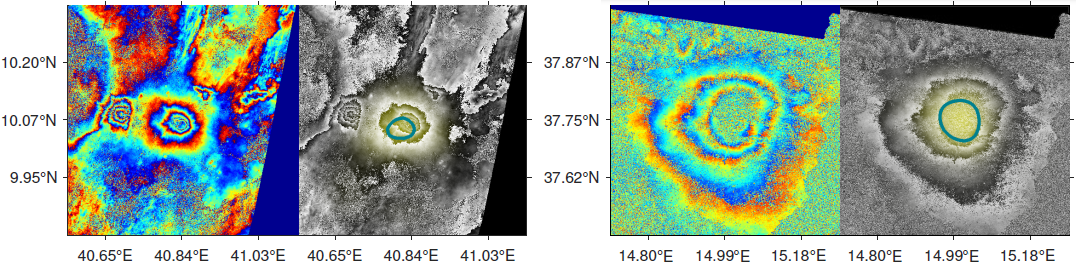}
				\includegraphics[width=\linewidth]{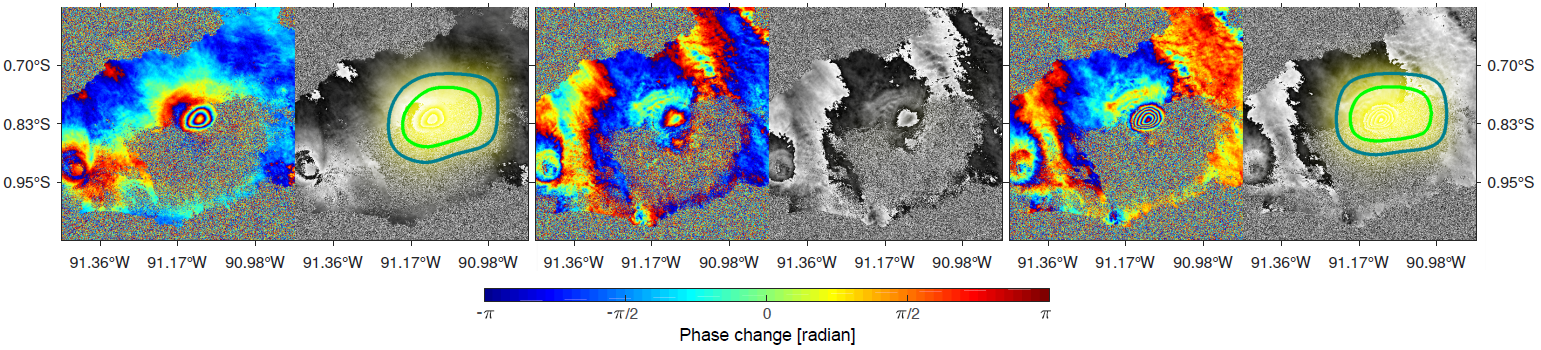}
	\caption{Results from retraining process using a combination of synthetic and real examples. Each pair shows the original image (left) and overlaid with probability of being volcanic deformation (right). Top row shows common false positives of all combination models, where the left plot is at Adwa (20170516-20170609) $P_{max}$=0.528 and the right plot is at Etna (20161214-20170302) $P_{max}$=0.547.  Bottom row shows three successive interferograms at Sierra Negra, where the left to the right plots are at (20170425-20170531), (20170519-20170531), (20170519-20170718), respectively. The middle plot was acquired from the shortest duration and the interferogram shows only one possible fringe, where the CNN models failed to detect this deformation, $P_{max}$=0.274. However,  the deformation signals were clearer in the longer-duration interferograms,  $P_{max}$=0.988 (left) and  $P_{max}$=1 (right). Areas inside dark and bright green contours are where $P>$0.5 and $P>$0.8, respectively. Each colour cycle (fringe) represents 2.8 cm of displacement in the satellite line-of sight.}
    \label{fig:retrainedresults}
\end{figure*}

{For valid evaluation, we carefully selected the false positive patches employed in the retraining process by spatially-shifting the test patches. This means none of the retraining data was the same as the test data.} The results of the combination models are shown in Table \ref{tab:Classification}. Following training with a combination of both synthetic and real examples, our framework achieves the best positive predictive value (PPV) of 82\% using the ``$\alpha D$+$\beta S$+$\gamma T$" model, followed by the ``$D$+$S$+$T$ vs $S$+$T$" model with a PPV of 79\%. These models reduce the number of false positives by more than half when compared to our previous study \citep{Anantrasirichai:Application:2018}, where the initial CNN model was trained using real data from the Envisat satellite.
However, there is one false negative at Sierra Negra (20170519-20170531) as shown in Fig. \ref{fig:retrainedresults} bottom row. This interferogram contains less than one fringe of deformation, and the expert only identified this as deformation because longer time period interferograms contain more fringes (e.g. interferograms of 20170425-20170531 and 20170519-20170718). These interferograms  were correctly identified by all combination models, and this illustrates the potential of using stacked data to identify smaller rates of deformation. Even the best-performing combination model still identified 9 false positives, examples of which are shown in Fig. \ref{fig:retrainedresults}. 

\subsection{Receiver Operating Characteristics}
\label{ssec:roc}
\begin{figure}[!th]
	\centering
      		\includegraphics[width=0.8\linewidth]{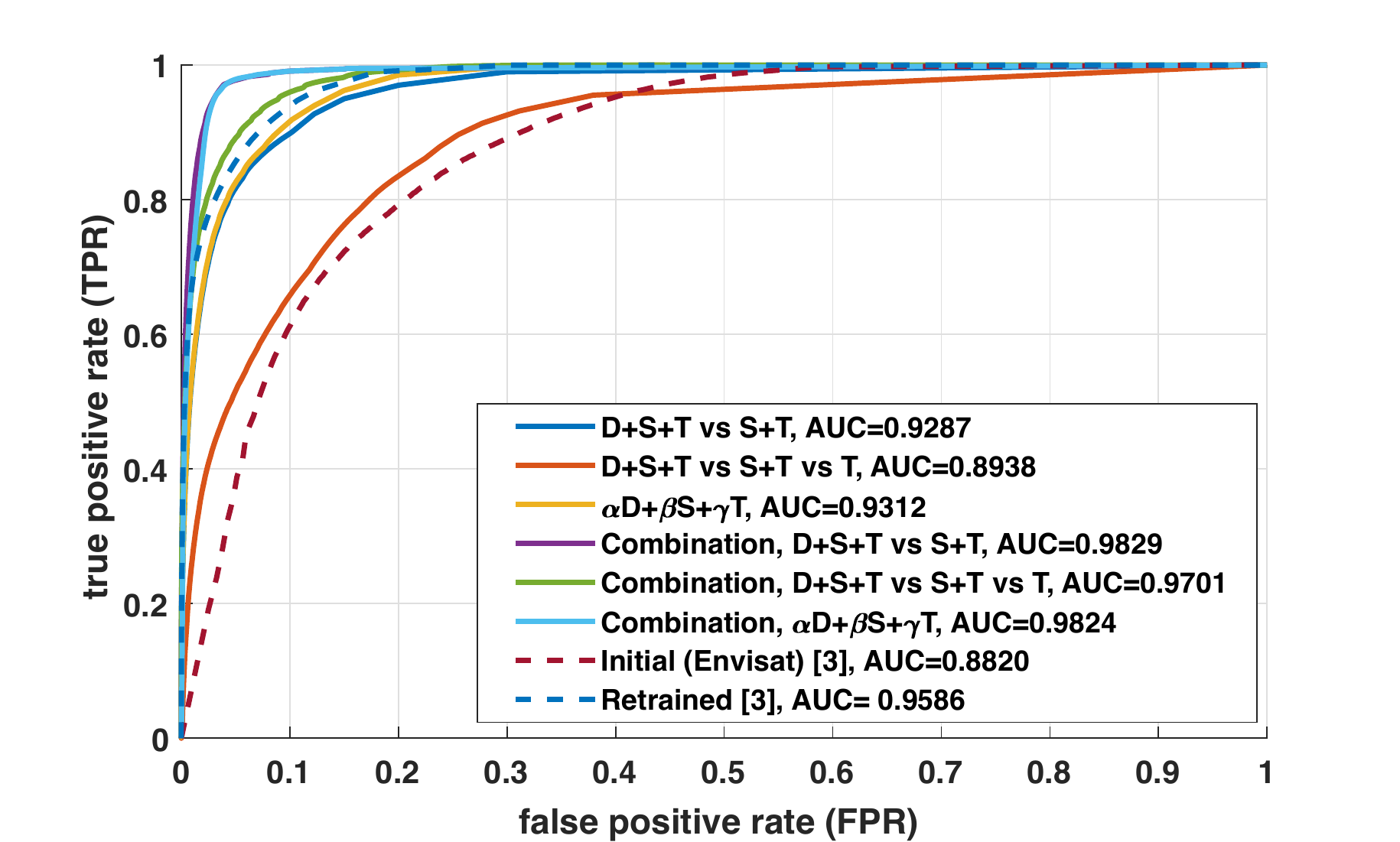}
	\caption{Receiver operating characteristic (ROC) curves for deformation detection on 1160 Sentinel-1 interferograms, which are all detected as positive by  ``$D$+$S$+$T$ vs $S$+$T$ vs $T$" model. These compare classification performances between the models of ``$D$+$S$+$T$ vs $S$+$T$", ``$D$+$S$+$T$ vs $S$+$T$ vs $T$", ``$\alpha D$+$\beta S$+$\gamma T$", and their combination models. The plot also includes the results of our previous work \cite{Anantrasirichai:Application:2018}, both the initial model with Envisat data and the retrained model. AUC = area under the curve.}
    \label{fig:ROCcurveR}
\end{figure}

The results were evaluated using a receiver operating characteristic (ROC) curve shown in Fig. \ref{fig:ROCcurveR}. This compares performance of each detection algorithm by calculating true positive (TPR) and false positive rates (FPR) by varying the probability thresholds for identify positives and negatives on the probability map following the Gaussian merge process. (see Fig. \ref{fig:workdiagram}). The TPR is the fraction of predicted positive samples that are retrieved over the total number of actual positive samples, whilst the FPR is the number of negative samples wrongly identified as positive divided by the total number of actual negative samples.
Fig. \ref{fig:ROCcurveR} also shows the area under curve (AUC), which is a metric for binary classification measured by integrating all area under the ROC curve.  Good classifiers will give high AUC values as they can detect the deformation signals correctly and few true negatives are falsely identified as deformation.  The ROC curve in Fig. \ref{fig:ROCcurveR} were computed using 1160 real interferograms predicted as positives by ``$D$+$S$+$T$ vs $S$+$T$ vs $T$" model. This provides a more useful performance comparison than using all 30,249 interferograms because the large number of correctly predicted negatives will give the AUC of close to 1 for all the models. The ROC curve reveals that the retraining process with both synthetic and real data improves the classification performance. The combination ``$D$+$S$+$T$ vs $S$+$T$" and ``$\alpha D$+$\beta S$+$\gamma T$ models outperform the others with the AUC of 0.983 and 0.982, respectively.
We also include the ROC curves of the initial model trained by Envisat data and its retrained model with Sentinel-1 data reported in our previous work \cite{Anantrasirichai:Application:2018}. The models trained with the synthetic samples outperform the initial model with Envisat data by up to 5.6\%, and the combination models can improve the performance by up to 2.5\% in term of the AUC. 

\section{Atmospheric correction }
\label{sec:atmcorrect}

The majority of the false positives contain strongly stratified atmospheric artefacts, which correlate with topography (Figure 4, Table \ref{tab:atmocorrection}). Although it is not feasible to apply atmospheric corrections to the entire dataset of $\>$30,000 inteferograms, doing so for the small number of positive detections is relatively simple. 

We manually apply the GACOS corrections as described in section \ref{ssec:Stratified} to each of the 51 positive results of the augmented ``$D$+$S$+$T$ vs $S$+$T$" model and an additional 2 false positives of the augmented ``$\alpha D$+$\beta S$+$\gamma T$". One true positive interferogram, Etna (20161003-20161015), was not processed because the the unwrapped file was not available from the LICSAR system. 
We request the GACOS zenithal tropospheric delays for the corresponding locations and acquisition dates and calculate the ZTD difference (slave-master) for each interferogram and reproject it in the corresponding line-of-sight.
For any missing values (incoherent regions), we interpolate them from the pixel values on the outer boundary of the missing regions \cite{Belhachmi:how:2009}. This technique computes the discrete Laplacian over the regions and  solves the Dirichlet boundary value problem to find a differential equation that is valid to the available outer boundary values.

\begin{figure*}[!ht]
	\centering
	\begin{tabular}{p{.02\textwidth}p{.35\textwidth}p{.5\textwidth}}
       &  \footnotesize Unwrapped interferograms  & \footnotesize Wrapped interferograms and detected results\\ 
    \end{tabular} 
	     \begin{tabular}{p{0.01\textwidth}p{0.15\textwidth}p{0.17\textwidth}p{.33\textwidth}p{.33\textwidth}}
       &  \footnotesize Uncorrected & \footnotesize Corrected & \footnotesize Uncorrected & \footnotesize Corrected \\ 
    \end{tabular} 
      \includegraphics[width=1\textwidth]{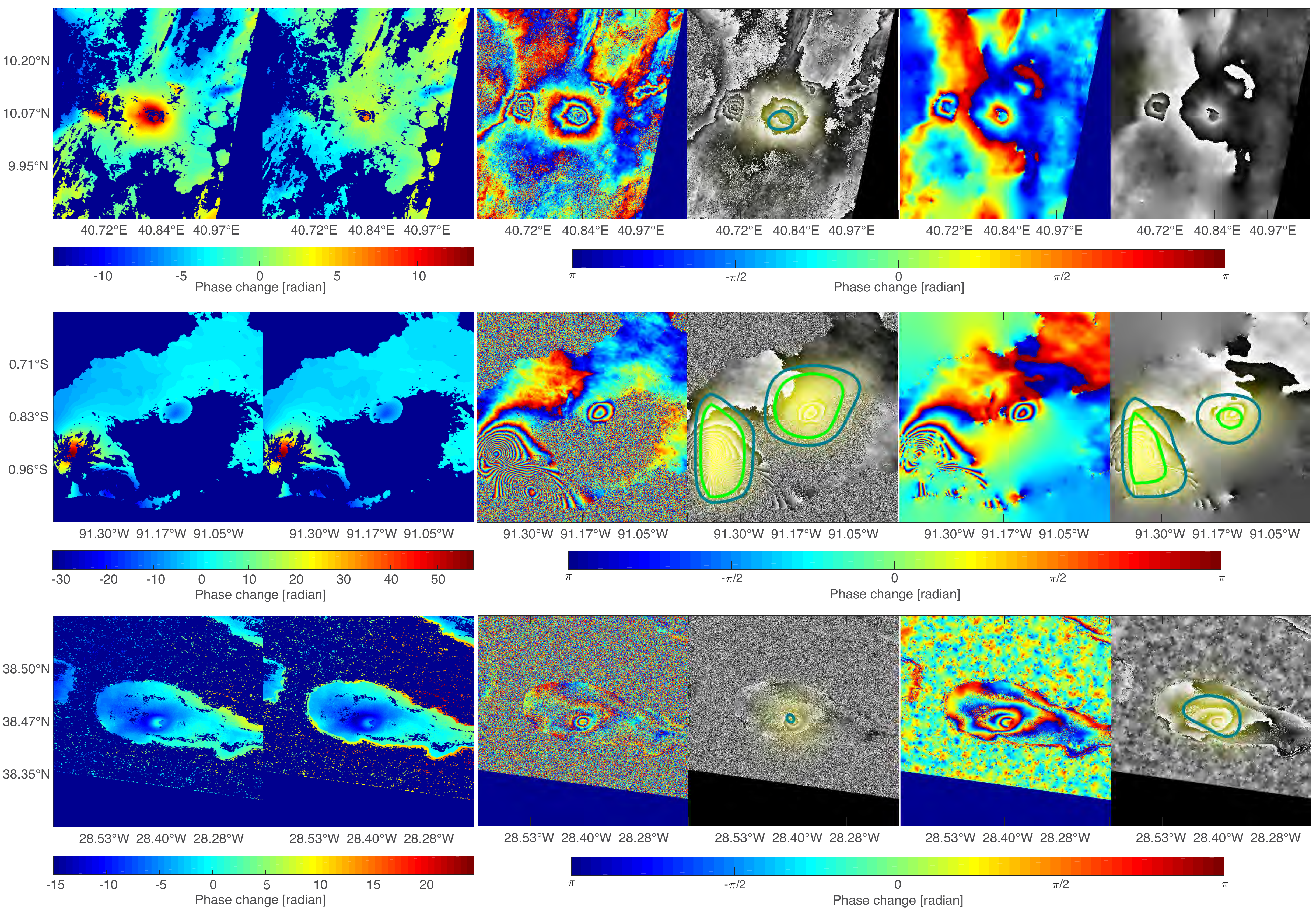} 
	\caption{Deformation detection results of the uncorrected and corrected interferograms. (top-row) Adwa (20170516-20170609) is an example of an improvement on false negative to true positive for all combination models. (middle-row) Sierra Negra (20170308-20170413) shows that the true positive result is still correctly identified for all combination models. (bottom-row) Pico (20170621-20170727) is the remaining false positive identified by only the combination ``$D$+$S$+$T$ vs $S$+$T$ vs $T$" model.
The brighter yellow means higher probability. Areas inside
dark and bright green contours are where $P>$0.5 and $P>$0.8, respectively.	Each colour cycle (fringe) of the wrapped interferograms represents 2.8 cm of displacement in the satellite line-of sight}
	\label{fig:all_corr_results}
\end{figure*}

 We then test the atmosphere-corrected inteferograms using the three best combination models, i.e. ``$D$+$S$+$T$ vs $S$+$T$" model, ``$D$+$S$+$T$ vs $S$+$T$ vs $T$" model, and ``$\alpha D$+$\beta S$+$\gamma T$" model. All 41 true positives are still identified as positives for all three combination models, which confirms that the atmospheric correction does not deteriorate the performance of the detection algorithm. 11 of 12 interferograms previously detected as false positives are now correctly identified by all three combination models. {One interferogram of Mount Pico (Fig. \ref{fig:all_corr_results} bottom row) does not contain deformation but is identified as positive by the ``$D$+$S$+$T$ vs $S$+$T$ vs $T$" model with $P_{max}$=0.652 after atmospheric correction. However, it is correctly identified by the ``$D$+$S$+$T$ vs $S$+$T$" model and $\alpha D$+$\beta S$+$\gamma T$ model with $P_{max}$=0.475 and 0.076, respectively.  Mount Pico is located on Pico Island in the Atlantic ocean and because the weather model resolution is approximately 10~km, the GACOS correction does not perform well on such a small island.}  Table \ref{tab:atmocorrection} shows the volcano list of false positive interferograms before applying atmospheric correction. The $P_{max}$ (maximum between the results of the combination ``$D$+$S$+$T$ vs $S$+$T$" and ``$\alpha D$+$\beta S$+$\gamma T$" models) shows the correct detection after mitigating atmospheric delay.

\begin{table*}
\caption{12 false positive interferograms before applying atmospheric correction from the combination ``$D$+$S$+$T$ vs $S$+$T$" and ``$\alpha D$+$\beta S$+$\gamma T$"  models. The $P_{max}$ of `uncorrected' and `corrected' interferograms are the maximum between the results of the combination ``$D$+$S$+$T$ vs $S$+$T$" and ``$\alpha D$+$\beta S$+$\gamma T$"  models. }
 \centering
\small
 \begin{tabular}{lccccc }
 \hline
 \multirow{ 2}{*}{Name} & \multirow{ 2}{*}{location} & \multirow{ 2}{*}{type} & \multirow{ 2}{*}{dates} & \multicolumn{2}{c}{$P_{max}$} \\
  &  &  &  & uncorrected & corrected \\ 
\hline
Adwa & Ethiopia & stratovolcano & 20170410-20170609 & 0.521 & 0.104 \\
Adwa & Ethiopia & stratovolcano & 20170516-20170609 & 0.528 & 0.001 \\
Alayta & Ethiopia & shield volcano & 20170104-20170305 & 0.512 & 0.000\\
Alayta & Ethiopia & shield volcano & 20170516-20170609 & 0.851 & 0.010 \\
Ale Bagu & Ethiopia & stratovolcano & 20170516-20170609 & 0.691 & 0.004 \\
Etna & Italy & stratovolcano & 20161027-20161202 & 0.689 & 0.001 \\ 
Etna & Italy & stratovolcano & 20161202-20161208 & 0.516 & 0.004 \\ 
Etna & Italy & stratovolcano & 20161214-20170302 & 0.547 & 0.045 \\ 
Etna & Italy & stratovolcano & 20170425-20170507 & 0.543 & 0.291 \\ 
Gran Canaria & Canary Islands & fissure vent & 20170417-20170423 & 0.626 & 0.465 \\
Gran Canaria & Canary Islands & fissure vent & 20170417-20170505 & 0.519 & 0.450 \\
Pico & Pico Island & stratovolcano & 20170621-20170727 & 0.507 & 0.475 \\
\hline
 \end{tabular}
 \label{tab:atmocorrection}
\end{table*}

Fig. \ref{fig:all_corr_results} compares the interferograms before and after applying the atmospheric correction. The first column is the unwrapped interferograms. The second and the third columns are the wrapped interferograms along with the classification results. The top row (Adwa (20170516-20170609)) reveals that the atmospheric correction can improve the classification performance, by reducing false positives. The second row (Sierra Negra (20170308-20170413)) shows correctly detected true positives. The third row (Pico (20170621-20170727)) shows the remaining false positive at Mt Pico where the weather model performs poorly.

\section{Discussion}
\label{sec:Discussion}

Satellite systems, such as InSAR, have the potential to routinely monitor surface deformation at volcanoes globally. They provide a large amount of data and make it available for public access. However, this has brought new challenges, as more data are impracticable to be analysed manually. Machine learning offers a possible solution, but the currently available Sentinel-1 dataset has a relatively small number of interferograms that show deformation, leading to a class imbalance problem in training datasets.

\subsection{Performance of the synthetically-trained CNN}
This paper presents a machine learning framework based on deep convolutional neural networks (CNNs) and solves the imbalanced training data using synthetic examples. It demonstrates the capability to identify rapid deformation signals from a large data set of interferograms with an improvement over our previous model which was trained using real interferograms  \citep{Anantrasirichai:Application:2018}. However, there are still some limitations in the current process, which are discussed in this section. These require further development before this could be used as an operational global alert system for volcanic unrest. 

The CNN models trained with the synthetic examples yield better generalisation performance, which is the ability of prediction on unseen samples, compared to the CNN models trained by real interferograms only. Using synthetic data overcomes the limitations caused by the small number of observed deformation signals, by including patterns which are physically plausible, but have not yet been observed.   However, the mathematical approximations that we use to generate our synthetic deformation signals simplify the volcano's magmatic plumbing system into a single source geometry (e.g. sphere, rectangular dislocation) and ignore heterogeneities in material properties and rheological parameters, but assuming an elastic half-space. Consequently, it cannot be guaranteed that our framework is sufficiently flexible to detect all possible deformation patterns. One possibility is to lower the probability threshold at which an image is flagged as `deformation' (currently we simply set the threshold at $P>$0.5), but this would increase the number of false positives (see Fig. \ref{fig:ROCcurveR}). Another possibility is outlier analysis, also known as anomaly detection, which can be used to identify rare events which differ significantly from the majority of the data \citep[e.g][]{Gaddam:kmeanID3:2007}. For this application, the background, including homogeneous areas, noise and atmospherically affected areas, could be modelled as a common multidimensional pattern \cite{Vlachos:Indexing:2006, Kieu:Outlier:2018}, and deformation flagged when the observed signal does not conform. 

Tropospheric delays in InSAR can be seperated into a turbulent component and a stratified component, which correlates with topography and often dominates the signal at volcanic edifices. As shown in Table \ref{tab:atmocorrection}, the false positives that remain after retraining the CNNs with a combination of real and synthetic data occur at the locations and times that have strong stratified atmospheric delays. In section \ref{sec:atmcorrect} we show that the atmospheric correction can be applied to improve detection accuracy. 
This however requires unwrapping algorithms which are computationally expensive, particularly in areas of low or patchy coherence. The correction method also relies on weather model, e.g. European Centre for Medium-Range Weather Forecasts (ECMWF) used in GACOS \cite{yu2018interferometric}, which is less accurate at the islands, where surrounding ocean causes complex and high tropospheric delays. However, by applying this correction only to positive results identified by the CNN, we reduce the overall computational expense without sacrificing accuracy. 

\subsection{Other sources of error}
The synthetic training dataset only considers very simple atmospheric conditions with a radially symmetric turbulent component, but more complex atmospheric phenomena are often observed in interferograms, e.g. atmospheric rolls, orographic effects (\cite[e.g.][]{Parker:Investigating:2014}). Similarly, our synthetic training dataset does not currently consider interferometric coherence.  Low coherence values are caused by changes to backscattering, and typically occur over densely vegetated areas, such as forests. Affected areas appear as random noise in the interferograms. Shadow areas cannot be reached by the radar pulses, and no backscatter is recorded at these locations.  These signals could be misinterpreted by our current models. To improve performance in these cases, the coherence map could be incorporated into the loss function (cost function), which is used for evaluating how well the learning algorithm models the given data. When calculating loss value during training, higher weights would be assigned to data with higher coherence as they are more reliable.

CNNs have proved their ability to capture noise characteristics through denoising applications \cite{Zhang:beyong:2017}. The trained kernels extract different low level features, such as brightness, lines and points. These features are combined in the higher layers to produce complex features and more semantic meaning. In this study, non-atmospheric noise is only learnt as part of the negative class during the retraining with real data, but synthetic examples could be included in future studies.

\section{Conclusions}
\label{sec:Conclusions}

This paper presents machine learning frameworks that automatically search through large volumes of wrapped InSAR images to detect rapid ground deformation that may be related to volcanic activity. The $>$30,000 short-term interferograms at over 900 volcanoes were systematically processed, but the majority of them were not deforming or were deforming slowly, leading to a problem of highly imbalanced training data. We solved this issue using synthetic examples, where three major components, i.e. deformation, stratified and turbulent atmospheres, were generated and combined for 2-, 3- and 91-class training. The synthetic deformations were generated using simple analytic models, the stratified atmospheres were acquired from the Generic Atmospheric Correction Online Service (GACOS), and the turbulent atmospheres were generated using the statistical properties of correlated noise. The classification models were then initialised with these synthetic datasets using the pretrained CNN, AlexNet. After an initial run, expert classification of the positive results were used to retrain the network with a combination of real and synthetic examples. The proposed framework achieves better performance than using the real interferograms alone -- reducing the number of interferograms that required manual inspection by half and decreasing the number of false positives by $>$80\%. Finally we present an atmospheric correction method used for analysing the suspicious positives. The results show that the combination CNN model can well classify the corrected interferograms.

\section{Acknowledgments}

This work was supported by the EPSRC Global Challenges Research Fund, the  NERC BGS Centre for Observation and Modelling of Earthquakes, Volcanoes and Tectonics (COMET), the NERC large grant Looking into the Continents from Space (NE/K010913/1), NERC innovation - Making Satellite Volcano Deformation Analysis Accessible (NE/S013970/1), the EPSRC Platform Grant - Vision for the Future (EP/M000885/1), and a seed grant from the University of Bristol Cabot Institute. The InSAR datasets are available at \url{https://volcanodeformation.blogs.ilrt.org/} and \url{http://comet.nerc.ac.uk/COMET-LiCS-portal/} and the training dataset is available at \url{https://seis.bristol.ac.uk/~eexna/download.html}.

\bibliography{machinelearning}

\begin{thebibliography}{10}
\expandafter\ifx\csname url\endcsname\relax
  \def\url#1{\texttt{#1}}\fi
\expandafter\ifx\csname urlprefix\endcsname\relax\def\urlprefix{URL }\fi
\expandafter\ifx\csname href\endcsname\relax
  \def\href#1#2{#2} \def\path#1{#1}\fi

\bibitem{Biggs:Global:2014}
J.~Biggs, S.~K. Ebmeier, W.~P. Aspinall, Z.~Lu, M.~E. Pritchard, R.~S.~J.
  Sparks, T.~A. Mather, Global link between deformation and volcanic eruption
  quantified by satellite imagery, Nature Communications 5 (2014) 3471--3471.
\newblock \href {http://dx.doi.org/10.1038/ncomms4471}
  {\path{doi:10.1038/ncomms4471}}.

\bibitem{Furtney:Synthesizing:2018}
M.~A. Furtney, M.~E. Pritchard, J.~Biggs, S.~A. Carn, S.~K. Ebmeier, J.~A. Jay,
  B.~T.~M. Kilbride, K.~A. Reath, Synthesizing multi-sensor, multi-satellite,
  multi-decadal datasets for global volcano monitoring, Journal of Volcanology
  and Geothermal Research 365 (2018) 38 -- 56.
\newblock \href {http://dx.doi.org/10.1016/j.jvolgeores.2018.10.002}
  {\path{doi:10.1016/j.jvolgeores.2018.10.002}}.

\bibitem{Parker:Systematic:2015}
A.~L. Parker, J.~Biggs, R.~J.Walters, S.~K.Ebmeier, T.~J.Wright, N.~A.Teanby,
  Z.~Lu, Systematic assessment of atmospheric uncertainties for insar data at
  volcanic arcs using large-scale atmospheric models: Application to the
  {C}ascade volcanoes, {U}nited {S}tates, Remote Sensing of Environment 170
  (2015) 102--114.
\newblock \href {http://dx.doi.org/10.1016/j.rse.2015.09.003}
  {\path{doi:10.1016/j.rse.2015.09.003}}.

\bibitem{Ebmeier:Applicability:2013}
S.~K. Ebmeier, J.~Biggs, T.~A. Mather, F.~Amelung, Applicability of {I}n{SAR}
  to tropical volcanoes: insights from {C}entral {A}merica, Geological Society,
  London, Special Publications 380 (2013) 15--37.
\newblock \href {http://dx.doi.org/10.1144/SP380.2}
  {\path{doi:10.1144/SP380.2}}.

\bibitem{pinel2014volcanology}
V.~Pinel, M.~P. Poland, A.~Hooper, Volcanology: lessons learned from synthetic
  aperture radar imagery, Journal of Volcanology and Geothermal Research 289
  (2014) 81--113.

\bibitem{bekaert2015statistical}
D.~Bekaert, R.~Walters, T.~Wright, A.~Hooper, D.~Parker, Statistical comparison
  of insar tropospheric correction techniques, Remote Sensing of Environment
  170 (2015) 40--47.
\newblock \href {http://dx.doi.org/10.1016/j.rse.2015.08.035}
  {\path{doi:10.1016/j.rse.2015.08.035}}.

\bibitem{li2005interferometric}
Z.~Li, J.-P. Muller, P.~Cross, E.~J. Fielding, Interferometric synthetic
  aperture radar (insar) atmospheric correction: Gps, moderate resolution
  imaging spectroradiometer (modis), and insar integration, Journal of
  Geophysical Research: Solid Earth 110~(B3).

\bibitem{jolivet2014improving}
R.~Jolivet, P.~S. Agram, N.~Y. Lin, M.~Simons, M.-P. Doin, G.~Peltzer, Z.~Li,
  Improving insar geodesy using global atmospheric models, Journal of
  Geophysical Research: Solid Earth 119~(3) (2014) 2324--2341.

\bibitem{ebmeier2016}
S.~Ebmeier, Application of independent component analysis to multitemporal
  insar data with volcanic case studies, Journal of Geophysical Research: Solid
  Earth 121~(12) (2016) 8970--8986.
\newblock \href {http://dx.doi.org/10.1002/2016JB013765}
  {\path{doi:10.1002/2016JB013765}}.

\bibitem{Gaddes2018}
M.~Gaddes, A.~Hooper, M.~Bagnardi, H.~Inman, F.~Albino, Blind signal separation
  methods for insar: The potential to automatically detect and monitor signals
  of volcanic deformation, Journal of Geophysical Research: Solid Earth
  123~(11) (2018) 10,226--10,251.
\newblock \href {http://dx.doi.org/10.1029/2018JB016210}
  {\path{doi:10.1029/2018JB016210}}.

\bibitem{Krizhevsky:ImageNet:2012}
A.~Krizhevsky, I.~Sutskever, G.~E. Hinton, Imagenet classification with deep
  convolutional neural networks, in: Proceedings of the 25th International
  Conference on Neural Information Processing Systems - Volume 1, Curran
  Associates Inc., USA, 2012, pp. 1097--1105.

\bibitem{Anantrasirichai:Application:2018}
N.~Anantrasirichai, J.~Biggs, F.~Albino, P.~Hill, D.~Bull, Application of
  machine learning to classification of volcanic deformation in
  routinely-generated insar data, Journal of Geophysical Research: Solid Earth
  123~(8) (2018) 6592--6606.
\newblock \href {http://dx.doi.org/10.1029/2018JB015911}
  {\path{doi:10.1029/2018JB015911}}.

\bibitem{Yan:deep:2015}
Y.~Yan, M.~Chen, M.~Shyu, S.~Chen, Deep learning for imbalanced multimedia data
  classification, in: IEEE International Symposium on Multimedia (ISM), 2015,
  pp. 483--488.
\newblock \href {http://dx.doi.org/10.1109/ISM.2015.126}
  {\path{doi:10.1109/ISM.2015.126}}.

\bibitem{Japkowicz:concept:2001}
N.~Japkowicz, Concept-learning in the presence of between-class and
  within-class imbalances, in: Advances in Artificial Intelligence, 2001, pp.
  67--77.
\newblock \href {http://dx.doi.org/10.1007/3-540-45153-6_7}
  {\path{doi:10.1007/3-540-45153-6_7}}.

\bibitem{Mogi:Relation:1958}
K.~Mogi, Relation between the eruptions of various volcanoes and deformations
  of the ground surfaces around them, Bull. Earthquake Res. Inst. Univ. Tokyo
  36 (1958) 99--134.

\bibitem{Okada:surface:1985}
Y.~Okada, Surface deformation due to shear and tensile faults in a half-space,
  Bulletin of the Seismological Society of America 75~(4) (1985) 1135.
\newblock \href {http://dx.doi.org/10.1016/0148-9062(86)90674-1}
  {\path{doi:10.1016/0148-9062(86)90674-1}}.

\bibitem{Fialko:Deformation:2001}
Y.~Fialko, Y.~Khazan, M.~Simons,
  \href{10.1046/j.1365-246X.2001.00452.x}{Deformation due to a pressurized
  horizontal circular crack in an elastic half-space, with applications to
  volcano geodesy}, Geophysical Journal International 146~(1) (2001) 181--190.
\newline\urlprefix\url{10.1046/j.1365-246X.2001.00452.x}

\bibitem{yu2018interferometric}
C.~Yu, Z.~Li, N.~T. Penna, Interferometric synthetic aperture radar atmospheric
  correction using a gps-based iterative tropospheric decomposition model,
  Remote Sensing of Environment 204 (2018) 109--121.
\newblock \href {http://dx.doi.org/10.1016/j.rse.2017.10.038}
  {\path{doi:10.1016/j.rse.2017.10.038}}.

\bibitem{Lohman:noise:2005}
R.~B. Lohman, M.~Simons, Some thoughts on the use of insar data to constrain
  models of surface deformation: Noise structure and data downsampling,
  Geochemistry, Geophysics, Geosystems 6~(1).
\newblock \href
  {http://arxiv.org/abs/https://agupubs.onlinelibrary.wiley.com/doi/pdf/10.1029/2004GC000841}
  {\path{arXiv:https://agupubs.onlinelibrary.wiley.com/doi/pdf/10.1029/2004GC000841}},
  \href {http://dx.doi.org/10.1029/2004GC000841}
  {\path{doi:10.1029/2004GC000841}}.

\bibitem{biggs2007multi}
J.~Biggs, T.~Wright, Z.~Lu, B.~Parsons, Multi-interferogram method for
  measuring interseismic deformation: Denali {F}ault, {A}laska, Geophysical
  Journal International 170~(3) (2007) 1165--1179.
\newblock \href {http://dx.doi.org/10.1111/j.1365-246X.2007.03415.x}
  {\path{doi:10.1111/j.1365-246X.2007.03415.x}}.

\bibitem{Matsugu:Subject:2003}
M.~Matsugu, K.~Mori, Y.~Mitari, Y.~Kaneda, Subject independent facial
  expression recognition with robust face detection using a convolutional
  neural network, Neural Networks 16~(5-6) (2003) 555--559.
\newblock \href {http://dx.doi.org/10.1016/S0893-6080(03)00115-1}
  {\path{doi:10.1016/S0893-6080(03)00115-1}}.

\bibitem{Li:CSRNet:2018}
Y.~Li, X.~Zhang, D.~Chen, {CSR}net: Dilated convolutional neural networks for
  understanding the highly congested scenes, in: 2018 IEEE/CVF Conference on
  Computer Vision and Pattern Recognition, 2018, pp. 1091--1100.
\newblock \href {http://dx.doi.org/10.1109/CVPR.2018.00120}
  {\path{doi:10.1109/CVPR.2018.00120}}.

\bibitem{Anantrasirichai:DefectNET:2019}
N.~Anantrasirichai, D.~Bull, Defect{NET}: multi-class fault detection on
  highly-imbalanced datasets, in: arXiv:1904.00863, 2019, pp. 1--5.

\bibitem{Agostinelli:Learning:2015}
F.~Agostinelli, M.~Hoffman, P.~Sadowski, P.~Baldi, Learning activation
  functions to improve deep neural networks, in: Proceedings of International
  Conference on Learning Representations, 2015, pp. 1--9.

\bibitem{He:Learning:2009}
H.~He, E.~A. Garcia, Learning from imbalanced data, IEEE Transactions on
  Knowledge and Data Engineering 21~(9) (2009) 1263--1284.
\newblock \href {http://dx.doi.org/10.1109/TKDE.2008.239}
  {\path{doi:10.1109/TKDE.2008.239}}.

\bibitem{ThaiNghe:cost:2010}
N.~Thai-Nghe, Z.~Gantner, L.~Schmidt-Thieme, Cost-sensitive learning methods
  for imbalanced data, in: The International Joint Conference on Neural
  Networks, 2010, pp. 1--8.
\newblock \href {http://dx.doi.org/10.1109/IJCNN.2010.5596486}
  {\path{doi:10.1109/IJCNN.2010.5596486}}.

\bibitem{Chawla:SMOTE:2002}
N.~V. Chawla, K.~W. Bowyer, L.~O. Hall, W.~P. Kegelmeyer, Smote: Synthetic
  minority over-sampling technique, Journal of Artificial Intelligence Research
  16 (2002) 321--357.
\newblock \href {http://dx.doi.org/10.1613/jair.953}
  {\path{doi:10.1613/jair.953}}.

\bibitem{Burgmann:Synthetic:2000}
R.~B\"{u}rgmann, P.~A. Rosen, E.~J. Fielding, Synthetic aperture radar
  interferometry to measure earth’s surface topography and its deformation,
  Annual Review of Earth and Planetary Sciences 28~(1) (2000) 169--209.
\newblock \href {http://dx.doi.org/10.1146/annurev.earth.28.1.169}
  {\path{doi:10.1146/annurev.earth.28.1.169}}.

\bibitem{Hanssen:radar:2001}
R.~Hanssen, Radar interferometry: Data interpretation and error analysis,
  Springer Netherlands, 2001.
\newblock \href {http://dx.doi.org/10.1007/0-306-47633-9}
  {\path{doi:10.1007/0-306-47633-9}}.

\bibitem{yu2017generation}
C.~Yu, N.~T. Penna, Z.~Li, Generation of real-time mode high-resolution water
  vapor fields from {GPS} observations, Journal of Geophysical Research:
  Atmospheres 122~(3) (2017) 2008--2025.

\bibitem{yu2018generic}
C.~Yu, Z.~Li, N.~T. Penna, P.~Crippa, Generic atmospheric correction model for
  {Interferometric Synthetic Aperture Radar} observations, Journal of
  Geophysical Research: Solid Earth.

\bibitem{Parsons:1994:2006}
B.~Parsons, T.~Wright, P.~Rowe, J.~Andrews, J.~Jackson, R.~Walker, The 1994
  sefiadbeh (eastern iran) earthquakes revisited: new evidence from satellite
  radar interferometry and carbonate dating about the growth of an active fold
  above a blind thrust fault, Geophys. J. Int 164~(1) (2006) 202--217.

\bibitem{Parker:Investigating:2014}
A.~Parker, J.~Biggs, Z.~Lu, Investigating long-term subsidence at medicine lake
  volcano, ca, using multitemporal insar, Geophysical Journal International
  199.
\newblock \href {http://dx.doi.org/10.1093/gji/ggu304}
  {\path{doi:10.1093/gji/ggu304}}.

\bibitem{Goodfellow-et-al-2016}
I.~Goodfellow, Y.~Bengio, A.~Courville, Deep Learning, MIT Press, 2016,
  http://www.deeplearningbook.org.

\bibitem{Canny:Edge:1986}
J.~Canny, A computational approach to edge detection, IEEE Trans. Pattern Anal.
  Mach. Intell. 8~(6) (1986) 679 -- 698.
\newblock \href {http://dx.doi.org/10.1109/TPAMI.1986.4767851}
  {\path{doi:10.1109/TPAMI.1986.4767851}}.

\bibitem{Belhachmi:how:2009}
Z.~Belhachmi, D.~Bucur, B.~Burgeth, J.~Weickert, How to choose interpolation
  data in images, SIAM J. Appl. Math. 70~(1) (2009) 333--352.
\newblock \href {http://dx.doi.org/https://doi.org/10.1137/080716396}
  {\path{doi:https://doi.org/10.1137/080716396}}.

\bibitem{Gaddam:kmeanID3:2007}
S.~R. Gaddam, V.~V. Phoha, K.~S. Balagani, K-{M}eans+{ID}3: A novel method for
  supervised anomaly detection by cascading k-means clustering and {ID}3
  decision tree learning methods, IEEE Transactions on Knowledge and Data
  Engineering 19~(3) (2007) 345--354.
\newblock \href {http://dx.doi.org/10.1109/TKDE.2007.44}
  {\path{doi:10.1109/TKDE.2007.44}}.

\bibitem{Vlachos:Indexing:2006}
M.~Vlachos, M.~Hadjieleftheriou, D.~Gunopulos, E.~Keogh, Indexing
  multidimensional time-series, The VLDB Journal 15~(1) (2006) 1--20.
\newblock \href {http://dx.doi.org/10.1007/s00778-004-0144-2}
  {\path{doi:10.1007/s00778-004-0144-2}}.

\bibitem{Kieu:Outlier:2018}
T.~Kieu, B.~Yang, C.~S. Jensen, Outlier detection for multidimensional time
  series using deep neural networks, in: IEEE International Conference on
  Mobile Data Management (MDM), 2018, pp. 125--134.

\bibitem{Zhang:beyong:2017}
K.~Zhang, W.~Zuo, Y.~Chen, D.~Meng, L.~Zhang, Beyond a gaussian denoiser:
  Residual learning of deep cnn for image denoising, IEEE Transactions on Image
  Processing 26~(7) (2017) 3142--3155.
\newblock \href {http://dx.doi.org/10.1109/TIP.2017.2662206}
  {\path{doi:10.1109/TIP.2017.2662206}}.

\end{thebibliography}

\end{document}